# A Differential Evolution-Enhanced Latent Factor Analysis Model for High-dimensional and Sparse Data

Jia Chen, *Member*, *IEEE*, Di Wu, *Member*, *IEEE*, and Xin Luo, *Senior Member*, *IEEE*

**Abstract**—High-dimensional and sparse (HiDS) matrices are frequently adopted to describe the complex relationships in various big data-related systems and applications. A Position-transitional Latent Factor Analysis (PLFA) model can accurately and efficiently represent an HiDS matrix. However, its involved latent factors are optimized by stochastic gradient descent with the specific gradient direction step-by-step, which may cause a suboptimal solution. To address this issue, this paper proposes a Sequential-Group-Differential- Evolution (SGDE) algorithm to refine the latent factors optimized by a PLFA model, thereby achieving a highly-accurate SGDE-PLFA model to HiDS matrices. As demonstrated by the experiments on four HiDS matrices, a SGDE-PLFA model outperforms the state-of-the-art models.

## 1. INTRODUCTION

HIGH-DIMENSIONAL AND SPARSE (HiDS) data are commonly adopted to store numerous incomplete yet useful information in various big data-related systems and applications[1-7]. These data describe the already known interactions between two high-dimensional entities and contain large amounts of latent knowledge[8-12]. Thereby, extracting latent knowledge from an HiDS matrix is vital in big-data-related analysis[13-17].

A Latent Factor Analysis (LFA) model is widely investigated and adopted to analyze an HiDS matrix[18-25]. To optimize an LFA model, the common approach is stochastic gradient descent (SGD) algorithm[1, 6, 18, 20] and some popular adaptive learning rate algorithms based on SGD. The representative algorithms contain AdaGrad[26], AdaError[27], AdaDelta[28] and Adam[29]. They adjust the instant learning rate by analyzing the past gradients in each iteration. Besides, recent studies propose to employ particle-swarm-optimization (PSO) to address these problems of the single searching path. ALF[30], HLFA[31], PLFA[32], and HPL[33] are the representative ones. However, the above algorithms have a common disadvantage that all the entities search optimum following specific paths[34-42], which limits the searching area.

Compared with these algorithms, Differential Evolution (DE) algorithm has a special divergence property[43-47]. In each evolution iteration, DE algorithm adopts the mutation and crossover strategies to perturb each entity, and then selects the optimal entities into the next iteration. This evolution scheme makes each entity mutate without relying on a continuous path. With this characteristic, DE is quite suitable for searching the area near optimal solutions. From this point of view, this study presents a Sequential Group DE-enhanced (SGDE) algorithm to refine the latent factors obtained by a PLFA model. This work makes the following contributions:

a) A SGDE-PLFA model is proposed. It consists of two layers. The initial layer adopts a PLFA model to pre-train the HiDS matrix, and the refinement layer performs the SGDE algorithm to search a more accurate optimum.
b) A SGDE algorithm is designed for refining a PLFA model. It divides the pre-trained latent factors into row sub-groups and column sub-groups. In each sub-group, an improved DE algorithm is adopted to refine this sub-group. All the LF sub-groups form the whole optimization process of an LFA model.

## 2. PRELIMINARIES

### 2.1 An LFA Model for HiDS Matrices

An HiDS matrix $R^{|U|\times|I|}$ represents the relationship between two large entities $U$ and $I$. Each element $r_{u,i}\in R$ represents a specific relationship value between $u\in U$ and $i\in I$. In $R$, $\Lambda$ represents the set of known entities. An objective function is defined as:

$$\varepsilon(P,Q) = \frac{1}{2}\sum_{r_{u,i}\in\Lambda}\left(r_{u,i}-\hat{r}_{u,i}\right)^2 = \frac{1}{2}\sum_{r_{u,i}\in\Lambda}\left(r_{u,i}-\sum_{k=1}^{f}p_{u,k}q_{k,i}\right)^2, \quad (1)$$

where $\hat{r}_{u,i}$, $p_{u,k}$ and $q_{k,i}$ represent the specific entries in matrices $\hat{R}$, $P$ and $Q$, respectively. According to [22], regularization terms and linear biases can be added into (1) to alleviate the magnitude effects in $R^{[48]}$. The objective function on a single instance $r_{u,i}$ can be represented as:

---

✧ Jia Chen is with the School of Cyber Science and Technology, Beihang University, Beijing 100191, China (e-mail: chenjia@buaa.edu.cn).
✧ Di Wu is with the Institute of Artificial Intelligence and Blockchain, Guangzhou University, Guangzhou 510006, Guangdong, China (e-mail: wudi.cigit@gmail.com)
✧ X. Luo is with the School of Computer Science and Technology, Chongqing University of Posts and Telecommunications, Chongqing 400065, China (e-mail: luoxin21@gmail.com).

$$\varepsilon_{u,i}(P,Q,\mathbf{b},\mathbf{c}) = \frac{1}{2}\left(r_{u,i} - \sum_{k=1}^{f} p_{u,k} q_{k,i} - b_u - c_i\right)^2 + \frac{\lambda}{2}\left(\sum_{k=1}^{f} p_{u,k}^2 + \sum_{k=1}^{f} q_{k,i}^2 + b_u^2 + c_i^2\right), \quad (2)$$

By optimizing (2) with SGD, we have the following scheme:

$$\forall r_{u,i} \in \Lambda, k \in (1,2,...,f): \begin{cases} p_{u,k}^\tau \leftarrow p_{u,k}^{\tau-1} - \eta \cdot \nabla \varepsilon_{u,i}\left(p_{u,k}^{\tau-1}\right), q_{k,i}^\tau \leftarrow q_{k,i}^{\tau-1} - \eta \cdot \nabla \varepsilon_{u,i}\left(q_{k,i}^{\tau-1}\right), \\ b_u^\tau \leftarrow b_u^{\tau-1} - \eta \cdot \nabla \varepsilon_{u,i}\left(b_u^{\tau-1}\right), c_i^\tau \leftarrow c_i^{\tau-1} - \eta \cdot \nabla \varepsilon_{u,i}\left(c_i^{\tau-1}\right), \end{cases} \quad (3)$$

where $\eta$ denotes the learning rate, $\nabla \varepsilon_{u,i}(\cdot)$ computes the stochastic gradient of $\varepsilon_{u,i}$, $\tau$ denotes the $\tau$-th iteration.

### 2.2 A DE Algorithm

The DE algorithm is a parallel direct search method that perturbs the entities with various mutation and crossover strategies, and then selects the proper entities for the next generation.

**Mutation**. With best/1 strategy, for each entity $\mathbf{d}_k$, the $k$-th mutation entity $\mathbf{dm}_k$ can be generated as:

$$\mathbf{dm}_k(\tau) = \breve{\mathbf{d}}(\tau-1) + m \cdot \left(\mathbf{d}_{rd1}(\tau-1) - \mathbf{d}_{rd2}(\tau-1)\right), \quad (4)$$

where $\breve{\mathbf{d}}(\tau\text{-}1)$ is the globally best entity until the $(\tau\text{-}1)$-th iteration, $rd_1, rd_2 \in \{1, 2, …, K\}$ are the random indexes, $m$ is the scaling factor of differential variation ($\mathbf{d}_{rd1}(\tau\text{-}1)\text{-}\mathbf{d}_{rd2}(\tau\text{-}1)$), respectively.

**Crossover**. DE mixes the sub-vectors of $\mathbf{dm}_k(\tau)$ with others. Two classical crossover strategies are the binomial (bin) and exponential (exp).

**Selection**. DE algorithm adopts fitness function to evaluate all the evolution entities. If an evolution entity yields a smaller fitness function value than the original one, it will be retained. The perturbation and selection processes make DE algorithm convergence fast without considering the gradient descent.

## 3. METHODOLOGY

This section emphasizes on the SGDE algorithm and an improved single sub-group DE optimization algorithm.

### 3.1 A Sequential-group DE Algorithm

The SGDE algorithm first maps all the latent factors into $(|U|+|I|)$ sub-groups, i.e., $|U|$ row vectors $[\mathbf{p}_u, b_u]$, $\forall u \in U$ and $|I|$ column vectors $[\mathbf{q}_i, c_i]$, $\forall i \in I$. Note that the SGDE algorithm sequentially initializes and optimizes the $(|U|+|I|)$ populations. We depict the sequential-group optimization process in Fig. 1.

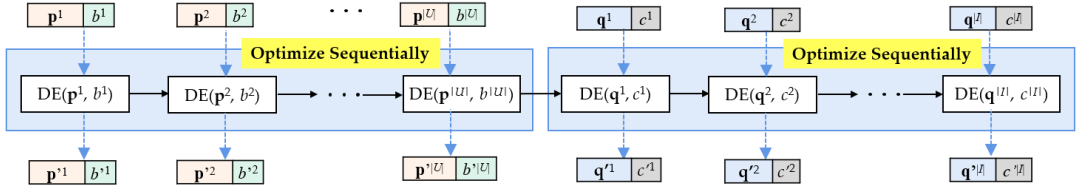

Fig. 1. A sequential-group DE Optimization for latent factors.

*3.1.1 Construction of LF-based DE populations*

First, we construct the $(|U|+|I|)$ sub-group LF-based DE populations via dividing the $\{P, Q, \mathbf{b}, \mathbf{c}\}$. Fig. 1 illustrates the detailed division of $\{P, Q, \mathbf{b}, \mathbf{c}\}$. The LF matrix of $[P, \mathbf{b}]$ consists of $|U|$ row vectors. Each row $[\mathbf{p}_u, b_u]$ in $R$, $\forall u \in U$ can be considered as an independent vector and randomly initialized to construct a DE population consisting of $K$ entities. Then, the $|I|$ rows in the LF matrix $[Q, \mathbf{c}]$ can be divided into $|I|$ independent vectors. Each $[\mathbf{q}_i, c_i]$, $\forall i \in I$ can construct a DE population consisting of $K$ vector entities.

More specifically, we fix the parameter set $\{\mathbf{p}_n \in P, Q, b_n \in \mathbf{b}, \mathbf{c}, \forall n \neq u\}$ and only optimize the $[\mathbf{p}_u, b_u]$, $\forall u \in U$. Each sub-vector $[\mathbf{p}_u, b_u]$ can be initialized randomly to form a $u$-th DE population consisting of $K$ vector entities.

In the $u$-th DE population constructed with $[\mathbf{p}_u, b_u]$, the first entity $\mathbf{d}_0^u$ is initialized as the value of $[\mathbf{p}_u, b_u]$. And

$$\mathbf{d}_k^u(0) = \begin{cases} [\mathbf{p}_u, b_u], k = 0, \\ \left[\mathbb{U}\left((1-\beta_p)\mathbf{p}_u, (1+\beta_p)\mathbf{p}_u\right), \mathbb{U}\left((1-\beta_b)b_u, (1+\beta_b)b_u\right)\right], k \in [1, K-1], \end{cases} \quad (6)$$

where $\mathbb{U}((1-\beta_p)\mathbf{p}_u, (1+\beta_p)\mathbf{p}_u)$ and $\mathbb{U}((1-\beta_b)b_u, (1+\beta_b)b_u)$ generate uniform distributed random numbers in the range of $[(1-\beta_p)\mathbf{p}_u, (1+\beta_p)\mathbf{p}_u]$ and $[(1-\beta_b)b_u, (1+\beta_b)b_u]$, respectively. With the $K$ initial entities in the $u$-th DE population, we calculate the globally best entity $\breve{\mathbf{d}}(0)$ at the 1st iteration. It can be updated as follows:

$$\breve{\mathbf{d}}(0) = \arg\min_{\mathbf{d}_k^u(0)}\left\{F\left(\mathbf{d}_k^u(0)\right)\right\}. \tag{7}$$

For $\forall i \in I$, we fix the parameter set $\{P, \mathbf{q}_n \in Q, \mathbf{b}, c_n \in \mathbf{c}, \forall n \neq i\}$ and only optimize the $i$-th sub-vector $[\mathbf{q}_i, c_i]$. Each sub-vector $[\mathbf{q}_i, c_i]$ can be initialized randomly to form a $i$-th DE population consisting of $K$ vector entities. In the $i$-th DE population, the first entity $\mathbf{d}_0^i$ is initialized as the value of $[\mathbf{q}_i, c_i]$. The other $\mathbf{d}_k^i$, $\forall k \in [1, K\text{-}1]$ can be initialized randomly in a small range around entity $\mathbf{d}_k^i$. The initialization formulas can be represented as:

$$\mathbf{d}_k^i(0) = \begin{cases} \left[\mathbf{q}_i, c_i\right], k=0, \\ \left[\mathbf{U}\left((1-\beta_q)\mathbf{q}_i,(1+\beta_q)\mathbf{q}_i\right), \mathbf{U}\left((1-\beta_c)c_i,(1+\beta_c)c_i\right)\right], k \in [1, K-1], \end{cases} \tag{8}$$

where $\mathbf{U}((1-\beta_q)\mathbf{q}_i, (1+\beta_q)\mathbf{q}_i)$ and $\mathbf{U}((1-\beta_c)c_i, (1+\beta_c)c_i)$ generate uniform distributed random numbers in the range of $[(1-\beta_q)\mathbf{q}_i, (1+\beta_q)\mathbf{q}_i]$ and $[(1-\beta_c)c_i, (1+\beta_c)c_i]$, respectively. With the $K$ initial values in the $i$-th DE population, we calculate the globally best entity $\breve{\mathbf{d}}(0)$ to share the global information. It can be updated as follows:

$$\breve{\mathbf{d}}(0) = \arg\min_{\mathbf{d}_k^i(0)}\left\{F\left(\mathbf{d}_k^i(0)\right)\right\}. \tag{9}$$

Based on the above division, the target LF set $\{P, Q, \mathbf{b}, \mathbf{c}\}$ is divided into $(|U|+|I|)$ sub-groups and initialized as $(|U|+|I|)$ DE populations with $K$ entities, whose initial values are near the corresponding sub-latent factor.

### 3.1.2 The improved single sub-group DE Algorithm

After initializing the $K$ entities, we perform an improved DE algorithm to refine each sub-latent factor group.

**Mutation.** For the DE populations initialized from the $|U|$ row vectors, each entity can be mutated with DE's mutation strategies. We illustrate the best/1 strategy here as an example. The $\mathbf{dm}_k^u(\tau)$ can be updated as:

$$\mathbf{dm}_k^u(\tau) = \breve{\mathbf{d}}(\tau-1) + m \cdot \left(\mathbf{d}_{rd1}^u(\tau-1) - \mathbf{d}_{rd2}^u(\tau-1)\right), \tau \geq 0 \tag{10}$$

where $\mathbf{d}_{rd1}^u(\tau\text{-}1)$ and $\mathbf{d}_{rd2}^u(\tau\text{-}1)$ are the randomly chosen entities, $m$ is the scaling factor, $\breve{\mathbf{d}}(\tau)$ is the best entity.

**Crossover.** Note that the crossover process is not performed in our improved DE algorithm. The reason is that each vector entity $u$ or $i$ in an HiDS matrix consists of $f$-dimensional latent factor attributes, which are dependent to each other. The crossover mixes some attributes from other entities, which breaks the dependency and integrality in a single latent factor vector. Thereby, we ignore crossover for the HiDS matrices processing.

**Fitness function.** We construct the fitness function to optimize parameter $[\mathbf{p}_u, b_u]$. The objective function in (2) can be re-constructed as the fitness functions:

$$F\left(\mathbf{d}_k^u(\tau)\right) = \frac{1}{2}\sum_{r_{u,i} \in \Lambda}\left(r_{u,i} - \mathbf{p}_u q_i - b_u - c_i\right)^2 + \frac{\lambda}{2}\left(\|\mathbf{p}_u\|^2 + |b_u|^2\right), \tag{11}$$

**Selection.** With (11), we calculate the fitness values of all the mutated entities $\mathbf{dm}_k^u(\tau)$, $\forall k \in K$, and perform the select process. The formula can be represented as:

$$\mathbf{d}_k^u(\tau) = \begin{cases} \mathbf{dm}_k^u(\tau), \text{ if } F\left(\mathbf{dm}_k^u(\tau)\right) < F\left(\mathbf{d}_k^u(\tau-1)\right), \\ \mathbf{d}_k^u(\tau-1), \text{ otherwise,} \end{cases} \tag{12}$$

Meanwhile, we calculate the best entity $\breve{\mathbf{d}}(\tau)$ to share the global information. Combining the above formulas, the single sub-group DE optimization for $[\mathbf{p}_u, b_u]$ performs iteratively till $\breve{\mathbf{d}}(\tau)$ converges or reaches the maximum iteration count. Note that, the similar optimization process is performed for $[\mathbf{q}_i, c_i]$, $\forall i \in I$. And with the above operations, we can optimize each single LF sub-group $[\mathbf{p}_u, b_u]$ or $[\mathbf{q}_i, c_i]$ via the improved DE algorithm. Then, the SGDE algorithm initializes and optimizes these DE populations sequentially.

## 4. PERFORMANCE ANALYSIS

### 4.1 General Settings

We choose Root mean squared error (RMSE) and mean absolute error (MAE) as the evaluation metrics[49]. We adopt four HiDS matrices from the industrial applications. Table I describes them. On the 4 datasets, we compare the proposed SGDE-PLFA model with the SGD- LFA, the Adam-LFA, and three PSO-SGD-LFA models ( PLFA, PLFA with linear biases, and HPL).

TABLE I. THE EXPERIMENTAL DATASETS AND THEIR DESCRIPTION

| No. | Dataset | $|U|$ | $|I|$ | $|\Lambda|$ | Data Density |
|---|---|---|---|---|---|
| D1 | ML10M | 10,681 | 71,567 | 10,000,054 | 1.31% |
| D2 | ExtEpinion | 775,760 | 120,492 | 13,668,320 | 0.015% |
| D3 | Flixter | 48,794 | 147,612 | 8,196,077 | 0.11% |
| D4 | Douban | 58,541 | 129,490 | 16,830,839 | 0.22% |

### 4.2 Performance Comparison

The comparison results of M1-6 in RMSE and MAE are represented in Table II. Fig.4 depicts the run time of all test models on D1-4. From the results, we have the following conclusions:

a) **SGDE-PLFA's achieves constantly higher prediction accuracy than that of its peers.** For example, as listed in Table II, M6's RMSE value is 0.8610 on D3, which is 8.72%, 9.02%, 8.69%, 1.26% and 0.53% lower than that of M1-5, respectively. All the other results are similar.

b) **SGDE-PLFA achieves significantly higher statistical accuracy than that of its peers.** M6 wins the SGD-LFA, the Adam-LFA, and three PSO-SGD-LF models on RMSE and MAE comparison. It obtains the lowest $F$-rank value among its peers. Furthermore, M6's Friedman testing scores in accuracy is 101.89, which is much higher than the critical value of 2.49 at the significance level of 0.05. That indicates M6 can predict the missing data with higher prediction accuracy than other models.

c) **M6 costs much less running time than the standard SGD-based and Adam-based LFA model.** For instance, compared with the running time on D2, M6 costs 700 seconds, which is 39.24% and 83.84% lower than M1's 1152 and M2's 4331 seconds, respectively. Besides, compared with the running time on D3, M6 costs 212 seconds, which is 52.99% and 93.75% lower than M1's 451 and M2's 3393 seconds, respectively.

d) **M5 and M6 cost comparable CPU running time with each other.** For instance, M5 costs 249 seconds on D1, which is a little smaller than M6's 280 seconds. While M5 costs 222 seconds on D3, which is a little higher than M6's 212 seconds.

Based on the three above findings, we conclude that an SGDE-PLFA model can achieve a significantly higher prediction accuracy than its peers. Furthermore, the DE-based refinement can search more accurate results than the PSO refinement with acceptable time consumption.

TABLE II. THE COMPARISON RESULTS ON RATING PREDICTION ACCURACY.

| Dataset | Metric | M1 | M2 | M3 | M4 | M5 | M6 |
|---|---|---|---|---|---|---|---|
| D1 | RMSE | 0.7872 | 0.7901 | 0.7858 | 0.7854 | 0.7850 | **0.7837** |
|  | MAE | 0.6069 | 0.6087 | 0.6046 | 0.6043 | 0.5998 | **0.5994** |
| D2 | RMSE | 0.7358 | 0.7342 | 0.7354 | 0.5361 | 0.5341 | **0.5292** |
|  | MAE | 0.3467 | 0.3451 | 0.3462 | 0.2983 | 0.2975 | **0.2577** |
| D3 | RMSE | 0.9433 | 0.9464 | 0.9429 | 0.8720 | 0.8656 | **0.8610** |
|  | MAE | 0.6653 | 0.6661 | 0.6641 | 0.6512 | 0.6394 | **0.6281** |
| D4 | RMSE | 0.7176 | 0.7213 | 0.7173 | 0.7076 | 0.7028 | **0.7027** |
|  | MAE | 0.5578 | 0.5582 | 0.5574 | 0.5567 | 0.5513 | **0.5496** |
| Statistic | Win/Loss | 8/0 | 8/0 | 8/0 | 8/0 | 8/0 | -- |
|  | $F$-rank* | 5.25 | 5.5 | 4.25 | 3 | 2 | **1** |

* A lower F-rank value indicates a higher rating prediction accuracy.

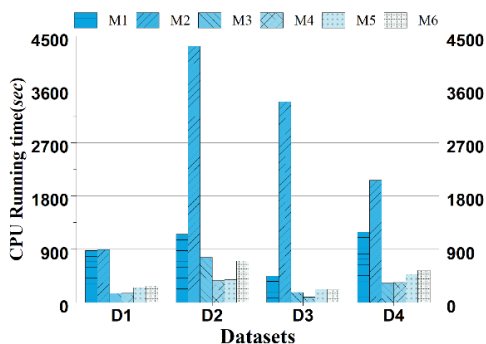

Fig. 2. CPU running time of M1-M6 on D1-4.

## 5. CONCLUSIONS

Based on the above experimental results and analyses, we conclude that a SGDE-PLFA model performs excellent in terms of prediction accuracy via refining the latent factors obtained by PLFA. And The additional time cost of the DE refinement is acceptable, and the efficiency of SGDE-PLFA is competitive with other related models.